# A Sparse Representation of Complete Local Binary Pattern Histogram for Human Face Recognition

Mawloud .Guermoui, Mohamed.L.Mekhalfi

*Abstract*— Human face recognition has been a long standing problem in computer vision and pattern recognition. Facial analysis can be viewed as a two-fold problem, namely (i) facial representation, and (ii) classification. So far, many face representations have been proposed, a well-known method is the Local Binary Pattern (LBP), which has witnessed a growing interest. In this respect, we treat in this paper the issues of face representation as well as classification in a novel manner. On the one hand, we use a variant to LBP, so-called Complete Local Binary Pattern (CLBP), which differs from the basic LBP by coding a given local region using a given central pixel and Sing_ Magnitude difference. Subsequently, most of LBP-based descriptors use a fixed grid to code a given facial image, which technique is, in most cases, not robust to pose variation and misalignment. To cope with such issue, a representative Multi-Resolution Histogram (MH) decomposition is adopted in our work. On the other hand, having the histograms of the considered images extracted, we exploit their sparsity to construct a so-called Sparse Representation Classifier (SRC) for further face classification. Experimental results have been conducted on ORL face database, and pointed out the superiority of our scheme over other popular state-of-the-art techniques.

## I. INTRODUCTION

Face recognition is a very important aspect in human surveillance. It has been a major focus in biometrics due to its non-invasive nature. Moreover, it is regarded as a primary method of person identification besides other alternative ways such as fingerprint. Many advantages can be gained behind human face recognition, the foremost leading ones trace back to security and identity verification purposes [1].

Human face recognition has known a fast growing interest, and become a very active topic in computer vision and biometrics. On the other side, it is particularly a challenging task given some key-issues related to lighting conditions, facial expressions, poses, orientations, image size and quality. Such problems may complicate the face recognition process.

In dealing with the above mentioned problems. Many face recognition techniques have been proposed during the past years. They can, generally, be categorized within two main branches, namely, (i) global, and (ii) local methods. Global methods make use of the whole face region as a raw input to a recognition system, such as for instance the well-known Principal Component Analysis (PCA) [2], Linear Discriminant Analysis (LDA) [3, 4], Independent Component Analysis (ICA) [5], Laplacianfaces [6], and the more recent 2D PCA [7]. Generally, such approaches can reveal good performance for classifying frontal views of faces. However, they are still vulnerable to a low efficiency when it comes to facial expressions, and pose changes. Furthermore, they are sensitive to face translation and rotation [8].

On the other hand, a number of local descriptors have also been proposed. Such methods can, in turn, be divided into two sub-categories, namely, sparse local descriptors, and dense local descriptors. Sparse local descriptors, which first detect the points of interest in a given image, and then extract a local patch and analyses it. Scale Invariant Feature Transform (SIFT) [9], is one of the classical sparse local algorithms proposed by David G. Lowe. It has been proven to have a good performance for object recognition, image retrieval, and other computer vision applications. Lately, SIFT features have also been adopted in many biometric applications especially for face recognition [10]. The main drawback of SIFT descriptor is its computational complexity and extensive number of extracted features (for a good performance), thus making it unsuitable for real-time applications.

Among the most popular dense local descriptors, which extract local information from a considered image, pixel by pixel, is the Local Binary Pattern (LBP). Originally introduced by Ojala et al [11, 12] for texture classification purposes. The main advantages of this operator are confined to its invariance to rotation, robustness against monotonic gray level transformation, and additionally its low computational complexity, which is a significant advantage over other former approaches.

The success of the LBP has inspired further studies in numerous applications, including face recognition [13, 14], facial expression recognition [15], face authentication [16], face detection [30], smart gun [17], fingerprint identification [18], automated cell phenotype image classification [19] and others. Since its advent [13], a lot of LBP variants have been proposed to overcome the shortcomings of the basic version. A comprehensive survey of different LBP variants can be found in [20].

In this paper, we propose to extract local information using Sign-Magnitude differences of CLBP [21] over multi-resolution decomposition of the given facial images. Compressive sensing [22-24], also known as sparse representation, which was proven as efficient in biometric

Mawloud. Guermoui and Mohamed. L. Mekhalfi, are with the Department of Electronics, Faculty of Technology, University of Batna, Batna 05000 , Algeria(E-mails:{gue.mouloud, mmedlamine, }@gmail.com).

recognition filed [25-27], has also been used in our work as a classification tool due to its proven prominent results.

The remainder of this paper is organized as follows. In section 2 we present the classical LBP operator and our proposed approach in details. Experimental results are reported in section 3. Section 4 concludes the paper and suggests future work.

## II. LOCAL BINARY PATTERN

### A. Classical Local Binary Pattern (LBP)

The LBP descriptor was first introduced by Ojala et al [11]. The original algorithm labels the pixels $f_p = (0,...,7)$ of a given image by thresholding each one of them, in the neighborhood of $3*3$ by the value of the central pixel $f_c$, and a binary code is accordingly produced $S(f_p - f_c)$ as given by Eq.(1),. See Fig.1 for an illustration of the basic LBP operator.

$$S(f_p - f_c) = \begin{cases} 1 & if \quad f_p \geq f_c \\ 0 & Otherwise \end{cases} \quad (1)$$

Then, by assigning a binomial factor $2^p$, the LBP code is computed as follows:

$$LBP = \sum_{p=0}^{7} S(f_p - f_c) 2^p \quad (2)$$

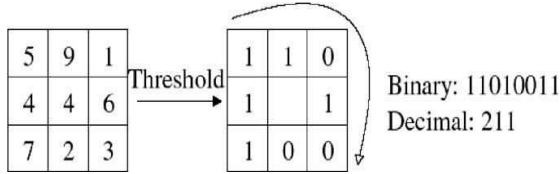

Figure1. The basic LBP operator.

Subsequently, this descriptor has afterwards been extended to other variants with the aim to capture discriminative features at different scales [12]. Therefore, a circular neighborhood around the central pixel $f_c$ is obtained using a bilinear interpolation, where the position of the neighborhood $f_p$ is determined by $[-R\sin(2\pi p/R), R\cos(2\pi p/R)]$. The subscript $LBP(P,R)$ denotes a neighborhood of P points of radius R with respect to the central pixel. This variant is commonly referred to as $LBP_{P,R}$.

Another important extension is the definition of "uniform patterns". An LBP is defined as uniform if it contains at most two *0-1* or *1-0* transitions when viewed as a circular bit string. To indicate the usage of two-transition uniform patterns, the superscript u2 is added to the LBP operator and uniform pattern are known as $LBP_{P,R}^{U2}$.

### B. Face description and recognition using LBP

The implementation of LBP descriptor for face representation is motivated by the fact that faces can be seen as a composition of micro patterns which are well described by such operator [13]. The idea consists in using the LBP descriptor to build several local descriptions of the face, and combine all of them into a global descriptor. The basic algorithm consists of dividing the face image into $m$ regions: $r_0, r_1, r_2 \ldots r_{m-1}$, the histograms of each region is computed using $LBP_{P,R}^{U2}$ then the resulting $m$ histograms are concatenated to obtain spatially enhanced histogram as illustrated in Fig. 2. The similarity distance between two faces is obtained using Chi square distance [13]:

$$X^2 = \sum_{i,j} \frac{(x_{i,j} - y_{i,j})^2}{x_{i,j} + y_{i,j}} \quad (3)$$

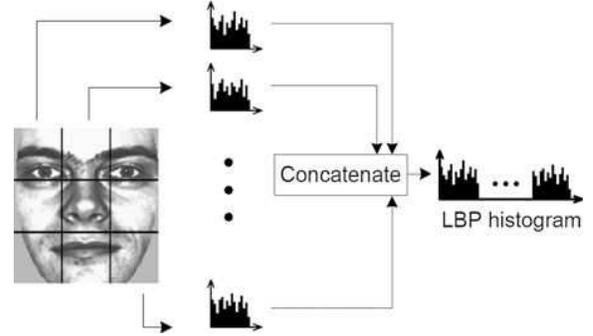

Figure 2. .Example of an LBP based facial representation [20].

### C. Complete Local Binary Pattern (CLBP)

In order to code the local image in a more complete manner, Guo et al, proposed CLBP pattern for texture classification. For a given pixel $f_c$ and its neighbors $f_p$, we calculate the difference between $f_p$ and $f_c$ : $D_p = f_p - f_c$. $D_p$ can be divided into the following parts:

$D_p = S_p * M_p$.

$$M_p = |D_p|, \quad S_p = \begin{cases} 1 & D_p \geq 0 \\ -1 & D_p \leq 0 \end{cases} \quad (4)$$

Where $S_p$ is the sign component and $M_p$ is the magnitude component. The CLBP pattern is defined using $f_c, M_p$ and $S_p$ separately and they are obtained as follow:

$$CLBP\_M_{P,R} = \sum_{p=0}^{p-1} t(M_p - c) 2^p.$$

$$t(x,c) = \begin{cases} 1, & x \geq c \\ 0, & x < c \end{cases} \quad (5)$$

Where c is the mean value of $M_p$ from the whole image.

$$CLBP\_S_{P,R} = \sum_{p=0}^{p-1} t(S_p, 0) 2^p \quad (6)$$

$$CLBP\_C_{P,R} = t(f_c - c_g) \quad (7)$$

Where $c_g$ is the mean value of the image.

The dimension of histogram corresponding to $CLBP\_M_{P,R}$ is $P+2$, the dimension of $CLBP\_S_{P,R}$ is also $P+2$ and for $CLBP\_C_{P,R}$ pattern is 2. An example of face image representation using CLBP pattern is illustrated in fig .3.

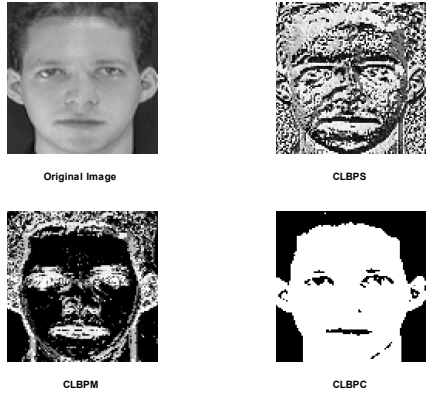

Figure 3. The CLBP pattern.

Simple concatenation of the $CLBP\_M_{P,R}$, $CLBP\_S_{P,R}$, and $CLBP\_C_{P,R}$ histograms gives better classification accuracy than the basic LBP operator.

Since local descriptors such LBP and CLBP assume that a given face region such as the eyes, the nose, etc. corresponds to the same face region over all the faces within the considered database, which is not always possible because they are often distributed among various areas like for instance the case of ORL database which cannot be characterized by a small size region. As a consequence, we use a multi-resolution grid (shown in fig.4) and for each sub-region of the pyramid representation we calculate CLBP_S and CLBP_M histogram separately and then concatenate the two histograms together, this scheme can be represented as "CLBP_S_M".

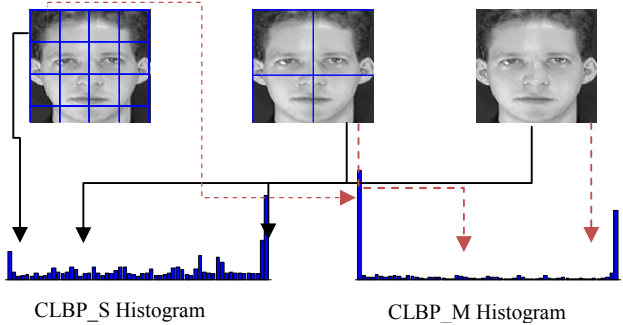

Figure 4. Multi-Resolution CLBP_S_M histogram construction.

After obtaining the CLBP_S_M histogram, sparse representation classifier is applied for recognizing different face classes. The sparse classifier is a suitable choice given that face images are, by their nature, sparse enough. We give the main classification procedure of SRC as used in [25]:

- Step1: We consider a dictionary matrix of training samples (i.e., consisting of the training image histograms stacked in one whole matrix) is divided into C classes $A = A_1 A_2 ... A_c \in R^{m*n}$, and a test sample $y_{test} \in R^m$. The dictionary matrix is formed by CLBP_S_M histogram.
- Step2: Normalize the columns of A to have $l^2$-norm and proceed by solving the $l^1$-minimization problem:

$$\hat{\alpha} = \arg\min \|\alpha\|_1 \quad subject\ to \quad A\alpha = y \quad (8)$$

- Step3: for each subject I, we compute the residual between the reconstructed sample $y_{recons}(i) = \sum_{j=1}^{n1} \hat{\alpha}_{i,j} y_{i,j}$ and the test sample by:

$$r(y_{recons}, i) = \|y_{k,test} - y_{recons}(i)\|_2 \quad (9)$$

- Step4: Identify the class of $y_{test}$ as:

$$y_{test} = \arg\min_i r(y_{tets}, i) \quad (10)$$

III. EXPERIMENTAL RESULTS

Proposed algorithm was assessed and compared to well-known state of the art techniques. For instance, experiments conducted on the ORL face database suggest that the proposed algorithm is more efficient as compared to PCA [2], LDA[4], LBP [13], SIFT [10], LGBPHs [28 ], and Gabor+LBP+LPQ [29].

A. ORL face database

ORL database contains ten different images of 40 distinct subjects. For some subjects, the images were taken at different times over a period of 2 years. There are variations in facial expression (open/closed eyes, smiling/nonsmiling.), facial details (glasses/no glasses) and head orientation. All the images were taken against a dark homogenous background with the subjects in an upright, frontal position, with tolerance for some tilting and rotation of up to about 20 degrees.

B. Evaluation

We select d images of each subject (d is the number of the first images of each subject) for training, while the remaining images are intended for testing purposes. To compare our method against other classic methods, we vary d from one image per subject to five images; the results of classification are illustrated in fig.5.

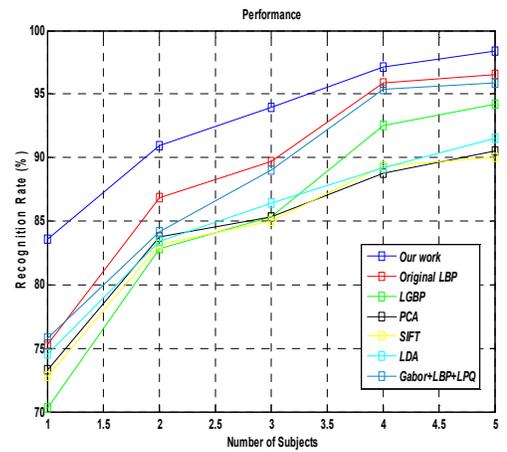

Figure 5. ORL –d (1, 2, 3, 4, 5) for gallery and others for probe.

Figure 5 shows the recognition rates regarding PCA based on Mahalanobis distance [2], LDA [4], basic LBP [13], LGBPHs [28], scale invariant feature transform (SIFT) [10],

Gabor+LBP+LPQ[29], and our proposed method. In this figure, the x coordinate stands for the number of training images per person, and the y coordinate points to the accuracy. It is to point out that the proposed method, for d=5, has yielded the highest accuracy (98.8%), followed by the basic LBP. It is also to say that the recognition rate increase gradually by increasing the training images.

In the second experiment, the reported accuracy is the averages over 10 runs, with 5 training images selected randomly and the remaining ones were exploited for test purposes. The mean recognition rate and standard deviation are given in table 1. In terms of both accuracy and stability, the proposed scheme outperforms the abovementioned state–of-the-art traditional methods, followed by the method proposed in [29].

| Method | ORL face database |
|---|---|
|  | Mean Recognition rate (%) |
| PCA[2] | 94.05±2.3387 |
| LDA[4] | 93.6±2.0385 |
| SIFT [10] | 93.70±1.23 |
| Original LBP [13] | 97.15± 1.203 |
| LGBPHS[28] | 95.60±1.8979 |
| Gabor+LBP+LPQ[29] | 95.5±1.29 |
| **Proposed approach** | **98.7 ±0.8756** |

Table.1. The recognition rates of the proposed method.

IV. CONCLUSION AND FUTURE WORK

In this work, a new scheme for face recognition is proposed. It makes use of a multi-resolution histogram of the well-known CLBP-S and CLBP-M for representing a given face image, and the compressive sensing technique as a sparse classifier. Proposed method was tested and compared versus different popular face recognition methods, the yielded experimental results revealed that the proposed method achieves higher recognition rate and better stability.

Future interests include more emphasis on the histograms extracted from the images as to combine them with other local descriptors. Another perspective related to the way in which we extract the most prominent bins of the feature vector without declining the recognition capability is also subject of interest.

REFERENCES

[1] Stan Z.Li.and Anil K .Jain, "Handboock of face recognition"Springer-Verlag London Limited 2011.
[2] M.Turk, A.Pentland, "Eigen faces for recognition," Journal of Cognitive Neuroscience 3(1991)72–86.
[3] P. Belhumeur, P. Hespanha, D. Kriegman, "Eigen faces vs. fisher faces: recognition using class specific linear projection," IEEE Transactions on Pattern Analysis and Machine Intelligence 19 (7) (1997) 711–720.
[4] K. Etemad, R. Chellappa, "Discriminant analysis for recognition of human face images," Journal of the Optical Society of America 14 (1997) 1724–1733.
[5] M.S. Bartlett, J.R. Movellan, T.J. Sejnowski , "Face recognition by independent component analysis, " IEEE Transaction on Neural Networks 13 (2002) 1450–1464.
[6] X. He, X. Yan, Y. Hu, P. Niyogi, and H. Zhang, "Face recognition using Laplacianfaces," IEEE Trans. Pattern Anal. Mach. Intell., vol. 27, no. 3, pp. 328–340.
[7] J. Yang, D. Zhang, A.F. Frangi, and J. Yang, "Two-dimensional PCA: A New Approach to Appearance-Based Face Representation and Recognition," IEEE Trans. Pattern Analysis and Machine Intelligence, vol. 26, no. 1, pp. 131-137, Jan.
[8] B. Heisele, P. Ho, J. Wu, and T. Poggio, " Face recognition: component-based versus global approaches," Compter Vision and Image Understanding, 91(1–2):6–21, 2003.
[9] D. Lowe, "Distince image features from scale-invariant key points. Int," Journal of Computer Vision, vol.60, no.2, pp.91- 110, 2004.
[10] M. Bicego, A. Lagorio, E. Grosso, and M. Tistarelli, " On the use of SIFT features for face authentication," Proc. of IEEE Int Workshop on Biometrics, in association with CVPR, NY, 2006.
[11] T. Ojala, M. Pietikainen, D. Harwood, " A comparative study of texture measures with classification based on featured distributions,"Pattern Recogn. 29(1) (1996) 51–59.
[12] T. Ojala, M. Pietikainen, and T. Mäenpää, " Multiresolution gray-scale and rotation invariant texture classification with local binary patterns," IEEE Trans. Pattern Anal. Mach. Intell., vol. 24, no. 7, pp. 971–987, Jul. 2002.
[13] Ahonen, T., Hadid, A., & Pietikainen, M, " Face recognition with local binary patterns," Proceedings of the European Conference on Computer Vision, LNCS 3021, pp.469-481, 2004.
[14] A.Hadid, M.Pietikäinen, S.Li, "Boosting spatio-temporal LBP patterns for face recognition from video," in: Proceedings of Asia-Pacific Workshop on Visual Information Processing, 2006, pp.75–80.
[15] C.Shan, S.Gong, P.W.McOwan, "Robust facial expression recognition using local binary patterns ,"in: Proceedings of IEEE International Conference on Image Processing, 2005, pp.914–917.
[16] Y.Rodriguez, S.Marcel, "Face authentication using adapted local binary pattern histograms," in: Proceedings of 9th European Conference on Computer Vision, 2006, pp.321–332.
[17] H.Jin, Q.Liu, H.Lu, X.Tong, "Face detection using improved LBP under Bayesian Framework," in: Proceedings of the Third International Conference on Image and Graphics, 2004, pp.306–309.
[18] Nanni L, Lumini A, " Local binary patterns for a hybrid fingerprint matcher," Pattern Recogn 2008; 11:3461–6.
[19] Nanni L, Lumini A, "A reliable method for cell phenotype image classification," Artif Intell Med 2008; 43(2):87–97.
[20] M. Pietikainen, A. Hadid, G. Zhao, and T, " Ahonen,Computer Vision Using Local Binary Patterns, "Springer-Verlag London Li-mited, 2011.
[21] Z. H. Guo, L. Zhang, and D. Zhang, " A Completed Modeling of Local Binary Pattern Operator for Texture Classification," IEEE Transactions on Image Processing, vol. 19, no. 6, pp. 1657-1663, Jun, 2010
[22] Donoho, D.L., "Compressed sensing," IEEE Trans. Information. Theory, vol. 52, no. 4, pp. 1289-1306. April, 2006.
[23] Baraniuk, R.G, "Compressive sensing," [lecture notes]. IEEE Signal Process. Mag. 2007, 24,118–121.
[24] Candes, E.J.; Wakin, M.B, "An introduction to compressive sampling," IEEE Signal Process. Mag. 2008, 25, 21–30.
[25] J.Wright, J.Yang, A.Y.Ganesh, A.Sastry, S.S. Yi Ma. , " Robust Face Recognition via Sparse Representation," IEEE Transactions on Pattern Analysis and Machine Intelligence, Vol.31,pp. 210 – 227, February 2009.
[26] S.Zhang, S.Zhao, B.Lei, " Robust Facial Expression Recognition via Compressive Sensing," Sensors 2012, 12, 3747-3761; doi: 10.3390/s120303747.
[27] C-H.Duan, C-K.Chiang, and S-H.Lai, " Face Verification with Local Sparse Representation," IEEE Signal Processing Letters, VOL. 20, NO. 2, 2013.
[28] W.Zhang, S.Shan, W.Gao, X.Chen, H.Zhang, " Local gabor binary pattern Histogram sequence (LGBPHS): a novel non-statistical model for face representation and recognition,". in: Proceedings of the Tenth IEEE International Conference on Computer Vision, 2005, pp.786–791.
[29] S-R.Zhou, J-P.Ying, and J-M.Zhang, " Local Binary Pattern (LPB) and Local Phase quantization (LPQ) Based on Gabor Filter for Face recognition ," Neurocomputing, 116 (2013), pp. 260–264.